\documentclass{bmvc2k}

\usepackage{graphics} 
\usepackage{epsfig} 
\usepackage{amsmath} 
\usepackage{amssymb} 
\usepackage{algorithm}
\usepackage{algpseudocode}
\usepackage{amsfonts}
\usepackage{mathtools}

\usepackage{times}
\usepackage{graphicx}
\usepackage{epstopdf}

\usepackage{multirow}
\usepackage{multicol}
\usepackage{adjustbox}
\usepackage{makecell}
\usepackage{booktabs}
\usepackage{here}

\usepackage[capitalise]{cleveref}

\title{Edge Detection for Event Cameras \\ using Intra-pixel-area Events}

\addauthor{Sangil Lee}{sangil07@snu.ac.kr}{1}
\addauthor{Haram Kim}{rlgkfka614@gmail.com}{1}
\addauthor{H. Jin Kim}{hjinkim@snu.ac.kr}{1}

\addinstitution{
Automation and Systems Research \\
Institute, Department of Mechanical \\
and Aerospace Engineering, \\
Seoul National University, \\
Seoul, Korea, Republic of
}

\runninghead{Lee, Kim, Kim}{Edge Detection for Event Cameras using Intra-pixel-area}

\def\etal{\emph{et al}\bmvaOneDot}

\begin{document}

\maketitle

\begin{abstract}
In this work, we propose an edge detection algorithm by estimating a lifetime of an event produced from dynamic vision sensor (DVS), also known as event camera. The event camera, unlike traditional CMOS camera, generates sparse event data at a pixel whose log-intensity changes. Due to this characteristic, theoretically, there is only one or no event at the specific time, which makes it difficult to grasp the world captured by the camera at a particular moment. In this work, we present an algorithm that keeps the event \textit{alive} until the corresponding event is generated in a nearby pixel so that the shape of an edge is preserved. Particularly, we consider a pixel area to fit a plane on Surface of Active Events (SAE) and call the point inside the pixel area closest to the plane as a intra-pixel-area event. These intra-pixel-area events help the fitting plane algorithm to estimate \textit{life time} robustly and precisely. Our algorithm performs better in terms of sharpness and similarity metric than the accumulation of events over fixed counts or time intervals, when compared with the existing edge detection algorithms, both qualitatively and quantitatively.
\end{abstract}

\section{Introduction}\label{sec:intro}

Event camera is a new type of vision sensor, which is motivated by the human eye, unlike standard CMOS cameras\cite{lichtsteiner2008128, brandli2014240}. Event camera is composed of an independent circuit for each pixel and these circuits generate \textit{events} asynchronously when the log-intensity of light applied to the pixel changes. These time-stamped event streams have small bit-rate of tens or more kilobytes, whereas the absolute intensity measurement of the standard camera has large bit-rate of tens or more megabytes. By these virtues, the event cameras have several advantages such as low latency, high temporal resolution, whereas the traditional camera captures image frame at a few tens of frames per second, usually. 

A wide range of fundamental computer vision applications, essential in autonomous navigation, object detection for avoidance, and augmented/virtual reality (AR/VR), can significantly benefit from the extremely low latency and fast response time of the event camera. However, most algorithms for computer vision cannot be applied directly to the event cameras due the new features mentioned above. So, there has been an increasing interest for developing algorithms suitable for event cameras in applications such as optical flow\cite{benosman2014event, gallego2018unifying}, visual odometry\cite{censi2014low, mueggler2014event, zhu2017event}, and SLAM\cite{kim2008simultaneous, kim2016real}. In this paper, we focus on the time-continuous edge detection algorithm for the event camera, which could be utilised for edge-based visual odometry\cite{jose2015realtime, kim2018edge, zhou2019canny} or SLAM\cite{maity2017edge}.

\begin{figure}[!t]
\begin{center}
\begin{tabular}{c c c c}
    \includegraphics[width=0.22\textwidth]{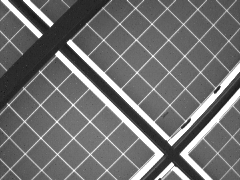} &
    \includegraphics[width=0.22\textwidth]{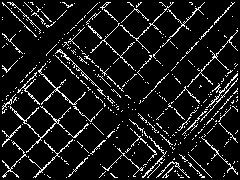} &
    \includegraphics[width=0.22\textwidth]{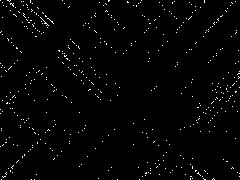} &
    \includegraphics[width=0.22\textwidth]{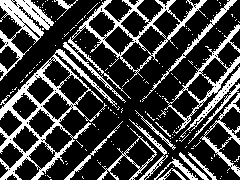} \\
    (a) Gray image & (b) Proposed & (c) 1 ms & (d) 30 ms \\
\end{tabular}
\end{center}
    \caption{The extracted sharp edge of the proposed algorithm.}
    \label{fig:introimage}
\end{figure}

The simplest way to detect an edge for the event camera is to accumulate events in a specified number or fixed time interval. However, its result is too sensitive to the speed of the camera or image because the value of accumulation volume is heuristically determined as shown in \cref{fig:introimage} (c) and (d). If the number of events generated per unit of time is less than the expected accumulation, the edge becomes sparse, otherwise, edge bleeding occurs. Also, such accumulation dilutes the advantages of an event camera (namely, low latency).

In the paper, we aim to detect thin edge pixels regardless of the speed of the camera and to fit local plane robustly against noise using intra-pixel-area pixel approach. Also, the proposed algorithm is designed based on an event-by-event basis, thus updates the edge pixels with low latency.

\subsection{Related Work}\label{ssec:work}

There are few methods developed for edge detection from the event camera data. Na\"ive methods are to accumulate events over a fixed time interval or numbers of events. However, in the case of accumulating events over time, the size of the interval depends on the speed of the camera because more events are generated when the camera moves faster. Therefore, the edge bleeding occurs or enough events are not triggered. On the other hands, if an edge is constructed by a fixed number of events accumulation, the quality of edge detection depends on the environment. A sequence captured in a simple environment with little gradient generates fewer events than in a complex environment. Consequently, it is difficult to determine a proper value of fixed time interval or number of events manually.

In order to detect edge robustly against various environmental properties, some algorithms have focused on detecting line edge in a structured environment. ELiSeD\cite{brandli2016elised} computes gradient direction of Surface of Active Events (SAE) with Sobel filters, and clusters similar orientations within neighbour events. However, due to the above accumulation approach, edge bleeding occurs, thus lines fitted on the clustered segments may become incorrect. In \cite{seifozzakerini2016event}, their event-by-event basis algorithm is inspired by the Hough transform and spiking neuron model. For each event, Hough transform converts their position to the so-called spikes in the $(r, \theta)$ space. The detected spikes stimulate the corresponding neuron in the $(r, \theta)$ space. This procedure is repeated while the potential value of the neuron exceeds the threshold, and the algorithm generates the line by using the triggered neuron. However, the performance depends on the resolution of $(r, \theta)$ parameter space, and is significantly degraded in areas where multiple lines meet.

Some research aim to detect edges, not just line segments that are frequently found in artifacts. F. Barranco \etal \cite{barranco2015contour} detects the contour of foreground objects. They extract features from the accumulated events such as orientation, timestamp, motion, and time texture. Then the boundary is predicted from the learned Structured Random Forest (SRF) given DVS features. However, since this algorithm is developed for object segmentation, it is prioritised to detect the boundary of the foreground, and the performance of the overall edge extraction may be degraded. E. Mueggler \etal \cite{mueggler2015lifetime} estimates the lifetime of event from local plane fitting on the SAE based on event-based visual flow\cite{benosman2014event}. However, na\"ive RANdom SAmple Consensus (RANSAC) method could not be successfully adapted for the event camera, thus causing imprecise estimation. Therefore, we propose a intra-pixel-area approach for RANSAC in order to estimate a local plane robustly and precisely. Also, we quantitatively evaluate algorithms in terms of similarity metric, which have not been done in most of the previous works.

\subsection{Contributions and Outline}\label{ssec:contrib}

Our main contributions can be summarised as follows: 
\vspace{-1mm}
\begin{enumerate}
	\item We propose a intra-pixel-area event approach for loss function of RANSAC, thus achieving robustness against noise and enhancing the performance of edge detection for event cameras.
	\vspace{-1mm}
	\item We evaluate the edge detection algorithms for the event camera by  quantitatively measuring similarity metric.
\end{enumerate}
\vspace{-1mm}
The rest of this paper is organised as follows: In \cref{sec:method}, we first characterise the event camera and describe the details of the algorithms. Next, we validate the performance of the proposed algorithm qualitatively in \cref{sec:exp}. Finally, \cref{sec:con} summarises the extension of this paper.

\section{Methodology}\label{sec:method}

First of all, we briefly review the definition of event cameras and the surface of active events in \cref{ssec:camera}. Next, we propose an event buffer pre-processing algorithm to remove noise events. In addition, the existing event-based visual flow and lifetime estimation by fitting a local plane on Surface of Active Events (SAE) is summarised, followed by the introduction of a intra-pixel-area event in \cref{sssec:floating}.

\subsection{Event Camera}\label{ssec:camera}

An event camera has independent pixels that detect light changes\cite{lichtsteiner2008128}. When the log-intensity of light applied to a pixel increases or decreases above/below the factory threshold, $\Delta \log(I) = \pm C$, the corresponding pixel $(x_i,y_i)$ generates an event ${\bf{e}}_i = (x_i,y_i,t_i,p_i)$ asynchronously. The event consists of time-stamp, $t_i$, $uv$-coordinated position of the pixel $(x_i,y_i)$, and polarity, $p_i$, where $1$ or positive means log-intensity increments and $0$ or negative means decrements.

In many existing algorithms for DVS\cite{benosman2014event,mueggler2015lifetime,mueggler2017fast}, the SAE is used to obtain information about when the event occurs in the corresponding pixel. When an event ${\bf{e}}_i$ appears, the timestamp of the event, $t_i$, is assigned to SAE($y_i$,$x_i$).

\subsection{Event Buffer}\label{ssec:buffer}

Even in the fixed event camera, the camera generates many events which are regarded as noise. These noise data can be suppressed by lowering the sensitivity of the event camera, but it also reduces the number of meaningful events that occur where the image intensity changes. To deal with noise events, we design an algorithm which precedes local plane fitting. The idea is that an event occurs along the gradient edge of the image, and the edge with adjacent pixels will move simultaneously on the image so that events will occur in a short time at adjacent pixels. In other words, events of the same polarity are generated during a short period of time within milliseconds along the same edge segment.

\begin{figure}[!t]
\begin{center}
\begin{tabular}{c c}
    \includegraphics[trim={0.5cm 0.2cm 0.8cm 0.4cm},clip,width=0.45\textwidth]{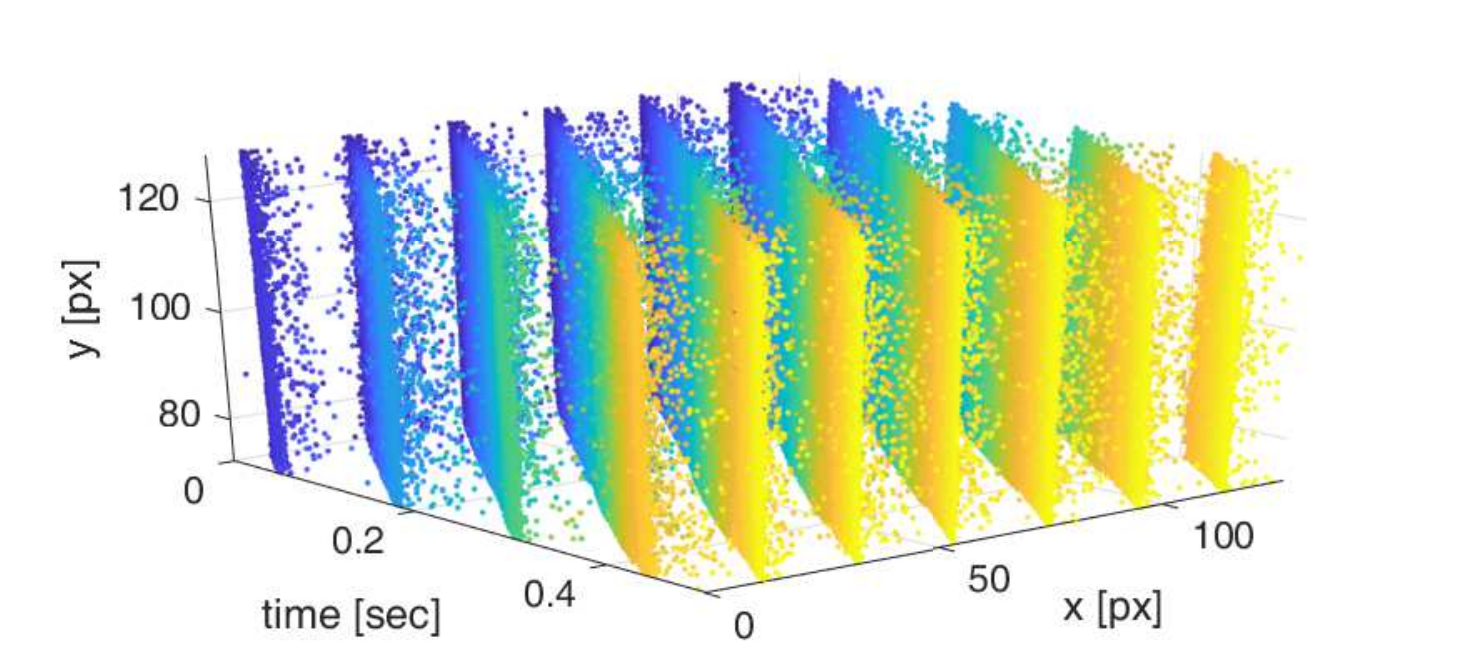} &
    \includegraphics[trim={0.5cm 0.2cm 0.8cm 0.4cm},clip,width=0.45\textwidth]{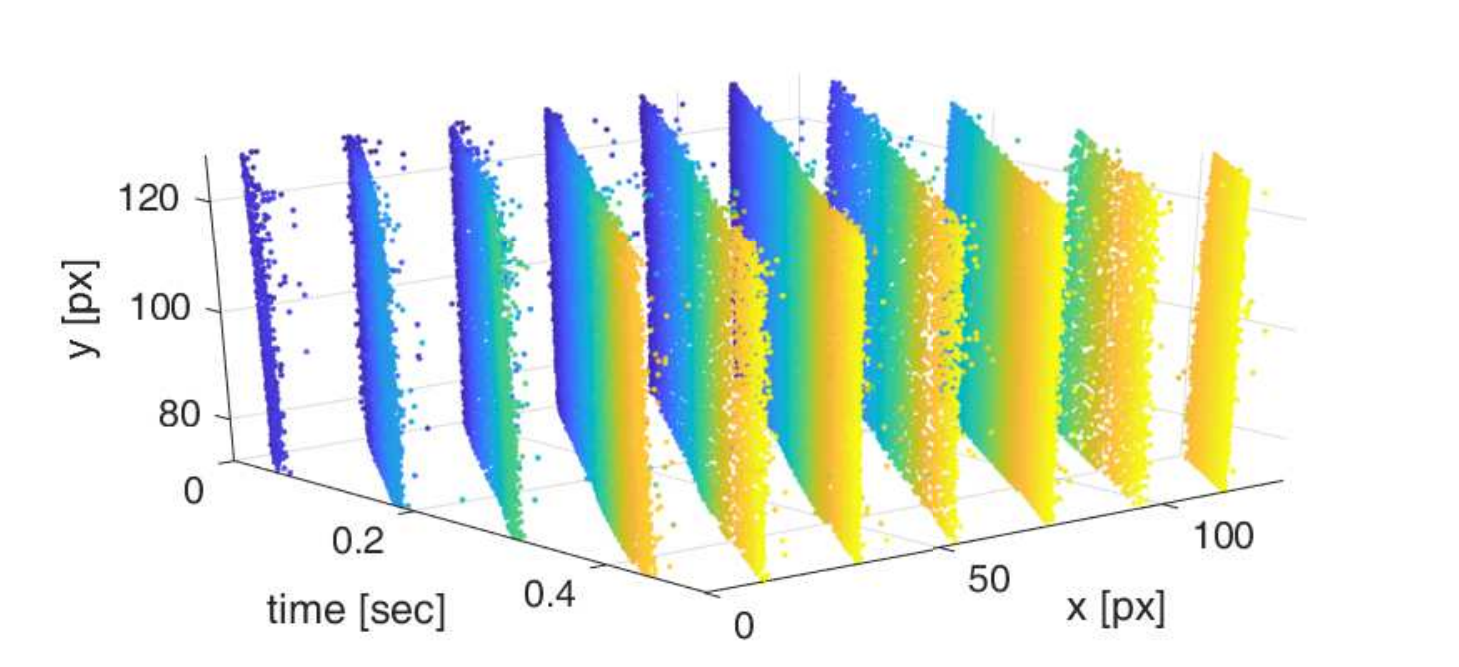} \\
    (a) Raw events & (b) The filtered events\\
\end{tabular}
\end{center}
    \caption{The $(x,y,t)$ accumulation of events for description of the event buffer. For the \texttt{stripes} sequence which is captured while the several lines move perpendicular to the camera principal axis. Between each plane, the noise appears as an isolated point.}
    \label{fig:eventbuffer}
\end{figure}

\cref{fig:eventbuffer} illustrates the SAE in the situation where the several straight stripes move perpendicular to the principle axis of the camera from $t=0$ to $t=0.5$. In the following evaluation including \cref{fig:eventbuffer}, we set $\tau_{min}=0.01$. As shown in \cref{fig:eventbuffer}(a), the raw event stream has a lot of noise data between planes. By pre-processing raw events with the event buffer procedure, we can reduce noise events successfully.

\subsection{Event-based Visual Flow and Lifetime}\label{ssec:fitting}

In order to estimate the lifetime of events, we utilise the existing algorithm\cite{benosman2014event,mueggler2015lifetime}. They suggest that the lifetime of an event means the time until the corresponding world point that has generated the current event before triggers a new event, and also this period indicates the maximum amount of time before a new event occurs nearby:
\begin{equation}
    \tau({\bf{x}}) = \max ~\Delta \hspace{.1em}t ~~ \textrm{subject to} ~~ ||{\bf{x}}||=1,
\end{equation}
where $\Delta \hspace{.1em}t = SAE({\bf{x}}+\Delta {\bf{x}})-SAE({\bf{x}})$ and ${\bf{x}} = (x,y)$ in pixel coordinates.

To find the steepest slope in the SAE, they estimate the normal vector of the plane, $\mathbf{n} = (n_1, n_2, n_3)$, which is fitted on the SAE locally, and visual flow and lifetime of the event are calculated below:
\begin{align}
    v_x &= -n_3/n_1, ~ v_y = -n_3/n_2, \\
    \tau({\bf{x}}) &= 1/v = 1/\sqrt{v_x^2+v_y^2} = \frac{1}{n_3} \sqrt{n_1^2 + n_2^2}.
\end{align}

\subsection{Local Plane Fitting}\label{sssec:localplane}

The event camera generates a \textit{digital} event when the average brightness of pixel changes, and the position of the generated event is fixed as the center point of the pixel area, thus the true position of an \textit{analog} event is not captured in the event stream. Therefore, we use RANSAC to robustly find a local plane fitted on the SAE. For RANSAC, we use past events in an $N \times N$ window including the current event which is the center of the window. Next, we use $k$-mean clustering to separate non-triggered or oldest event pixels from the $N^2$ pixels around the current event by setting $k$ as 2. The initial cluster centroid are set to the minimum and the current timestamp. The set including the current event is selected. Then, we choose the current event and two past events randomly in the selected set to compute the candidate plane.

\subsubsection{Intra-pixel-area Event}\label{sssec:floating}

By the way, the gradient change occurs at some sub-pixel point inside the pixel theoretically, but actually, the event is triggered by the change of the average brightness of a pixel, and the position of the event indicates the center of the pixel. Therefore, we introduce a \textit{intra-pixel-area event}, $S_{\bf{x}}(\delta)$, which exists somewhere inside a pixel area:
\begin{align}\label{eq:neighbour}
    S_{\bf{x}}(\delta) = \{(x,y)|x-\delta < {\bf{x}}_1 < x+\delta, y-\delta < {\bf{x}}_2 < y+\delta\},
\end{align}
where ${\bf{x}} = ({\bf{x}}_1,{\bf{x}}_2)$ is the pixel of the current event. Thus, we choose the current event and float event of two past events randomly in RANSAC. Then, we compute loss function as the distance between candidate plane and intra-pixel-area events, $\bf{z}$, and the distance function is defined as below:
\begin{align}
    d = \min \textrm{dist}(plane, \bf{z}),~ \forall \bf{z} \in S_{\bf{x}}(\delta),
\end{align}
where 
\begin{equation}
    \textrm{dist}(plane, {\bf{z}}) = \frac{|{\bf{n}}^{\rm{T}} {\bf{z}} - 1|}{||{\bf{n}}||^2_2} ~~\text{with a plane:}~~ {\bf{n}}^{\rm{T}} {\bf{x}} = 1
\end{equation}
which contains the current event.

\begin{figure}[!t]
\begin{center}
\begin{tabular}{c c c}
    \includegraphics[trim={1cm 0.5cm 0.9cm 0.5cm},clip,width=0.3\textwidth]{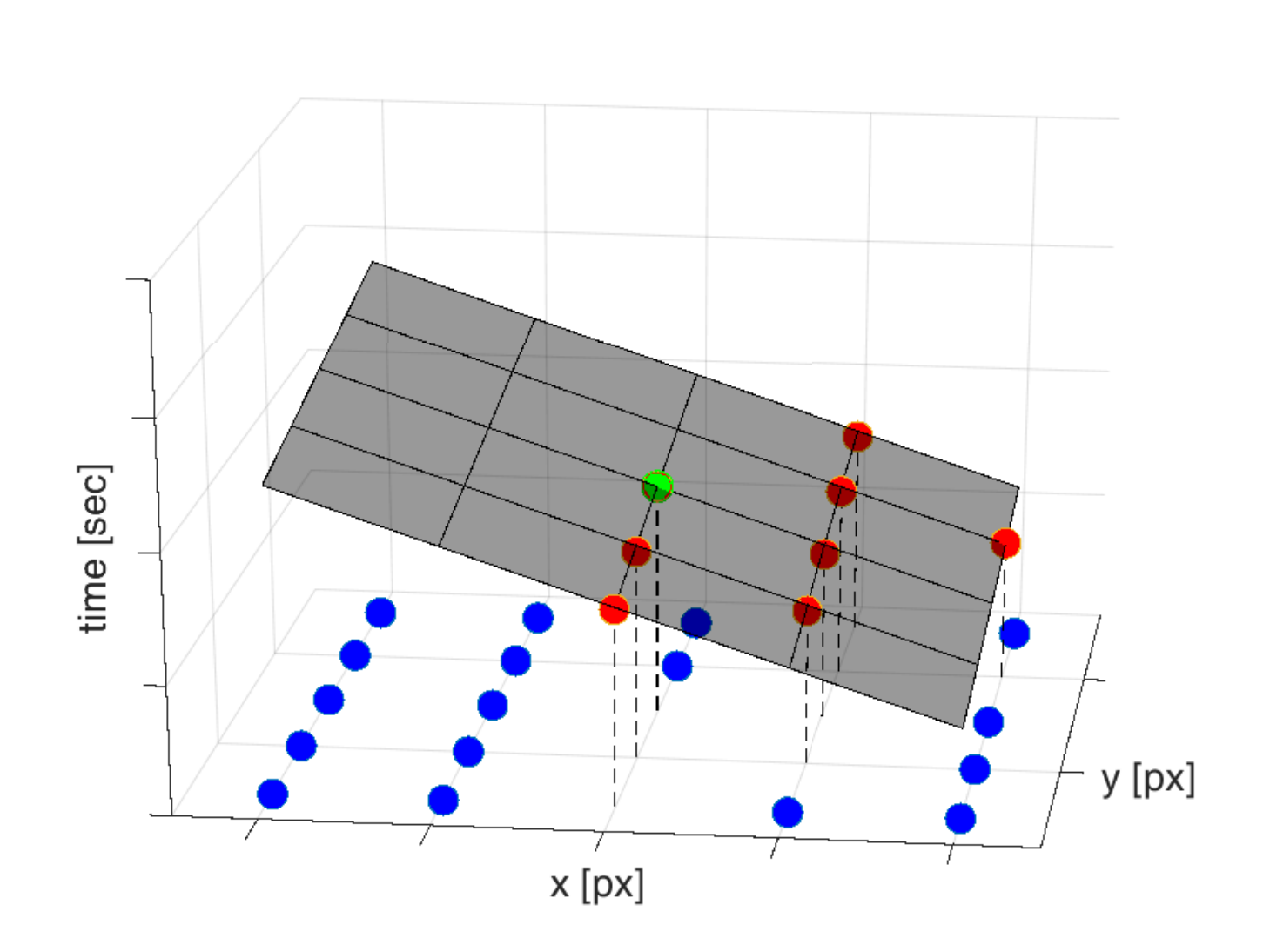} &
    \includegraphics[trim={1cm 0.5cm 0.9cm 0.5cm},clip,width=0.3\textwidth]{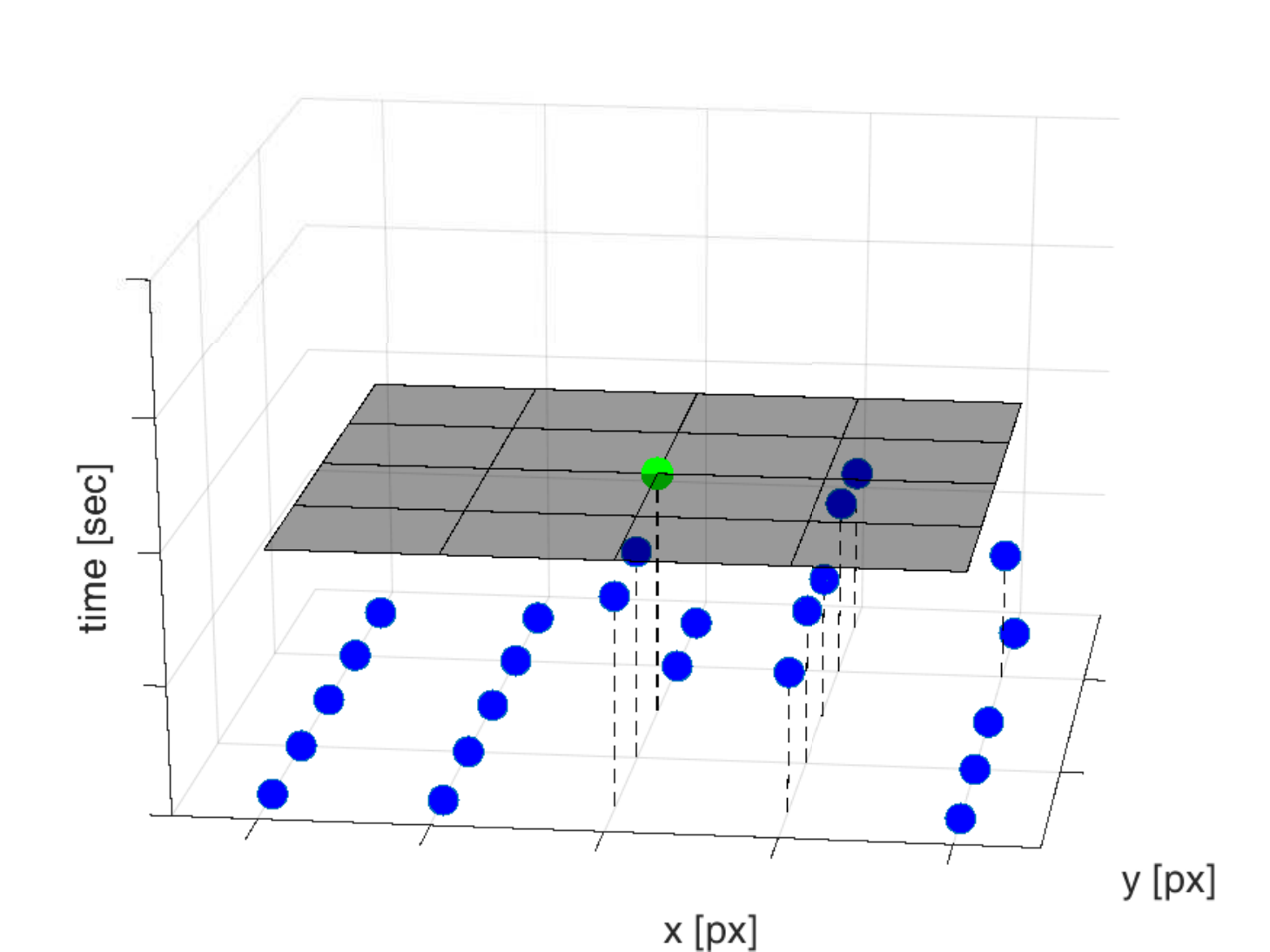} &
    \includegraphics[trim={1cm 0.5cm 0.9cm 0.5cm},clip,width=0.3\textwidth]{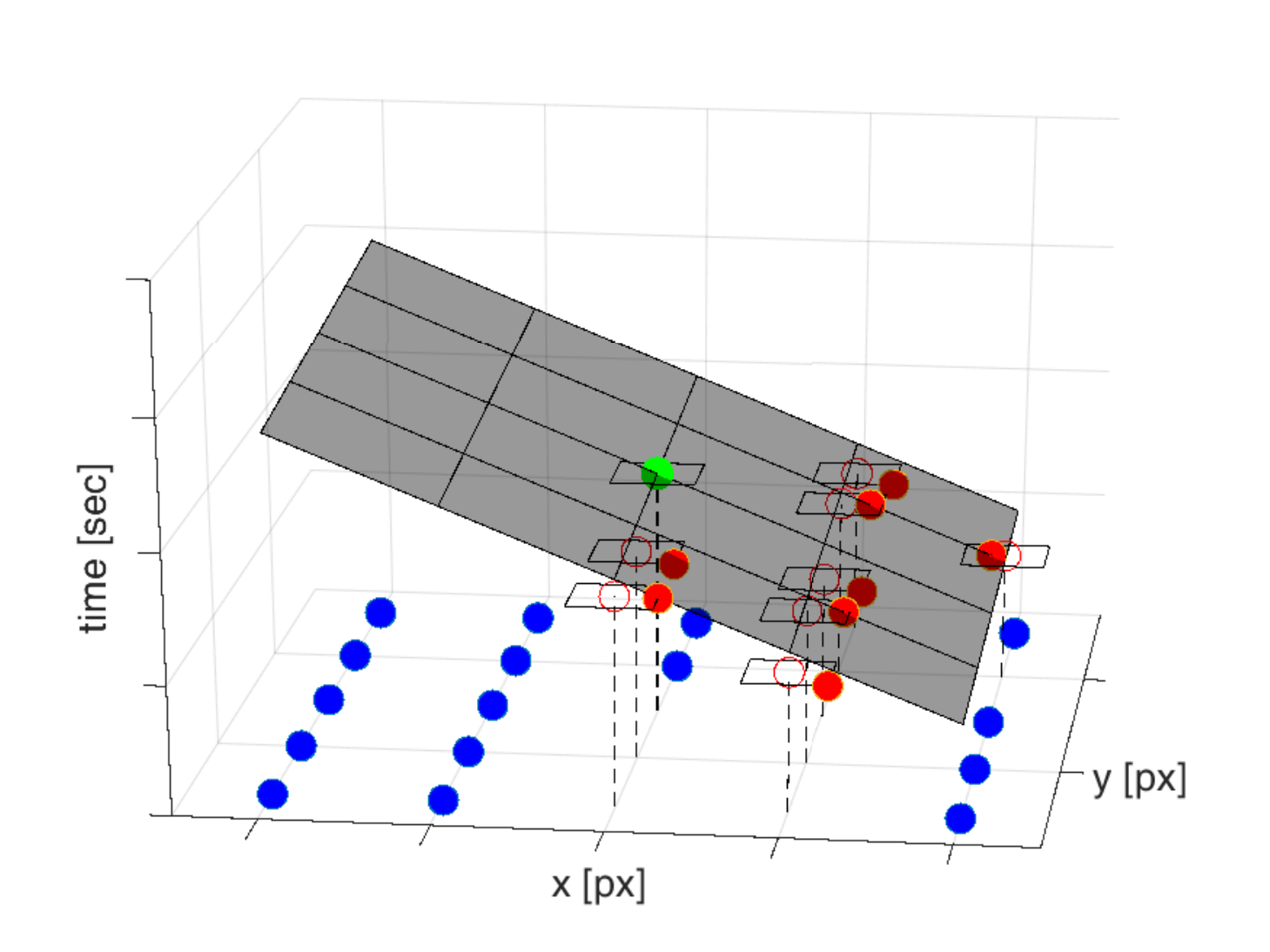} \\
    \makecell{(a) w/o intra-pixel-area \\under mild noise} &
    \makecell{(b) w/o intra-pixel-area \\under severe noise} & 
    \makecell{(c) w/ intra-pixel-area \\under severe noise}\\
\end{tabular}
\end{center}
    \caption{Description of the intra-pixel-area event. After computing a local plane (gray) with or without intra-pixel-area event by RANSAC, the outliers (blue), inliers (red), and the current event (green) are drawn. In (c), raw inlier events are represented as an \textit{unfilled} red circle, whereas the closest points in each intra-pixel area of inlier event are depicted as a \textit{filled} circle. Also, the intra-pixel areas of the inlier events are depicted as a rectangle.}
    \label{fig:floatingevent}
\end{figure}

We show the result of intra-pixel-area events in \cref{fig:floatingevent}. It describes the estimation of local plane fitted on the SAE around the current event (green). RANSAC is executed from events in a $5 \times 5$ window, and the value of events are given manually for simulation. In particular, the timestamp value of events in \cref{fig:floatingevent} (b) and (c) are contaminated by severe noise. Comparing the result of \cref{fig:floatingevent} (a), intra-pixel-area event approach makes the candidate plane more likely to be voted on by more events, taking into account the positional variation of past events, thus making more robust against the noise. 

\begin{figure}[!t]
\begin{center}
\begin{tabular}{c c}
    \includegraphics[trim={1cm 0.1cm 0.9cm 0cm},width=0.44\textwidth]{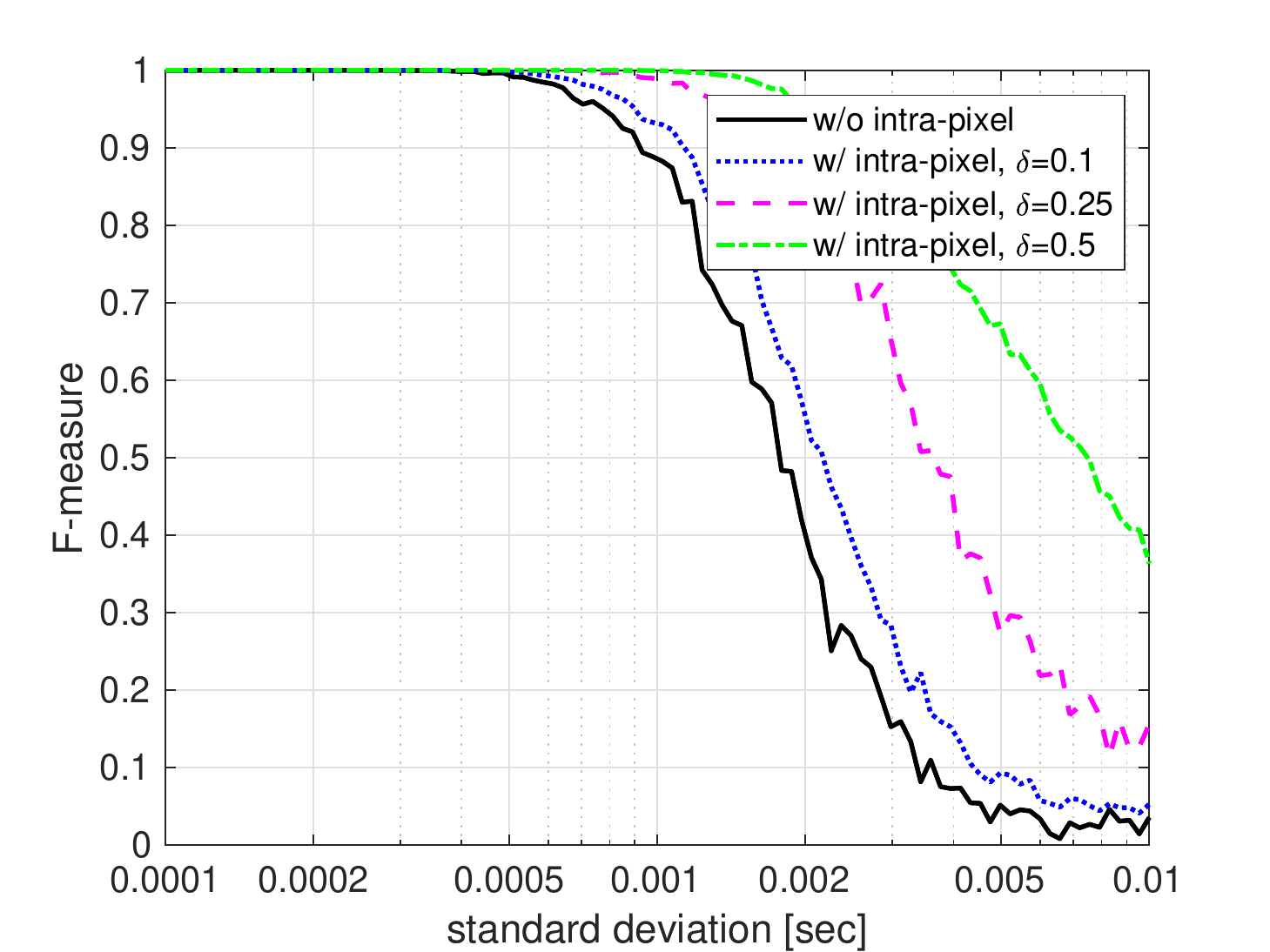} &
    \includegraphics[trim={1cm 0.1cm 0.9cm 0cm},width=0.44\textwidth]{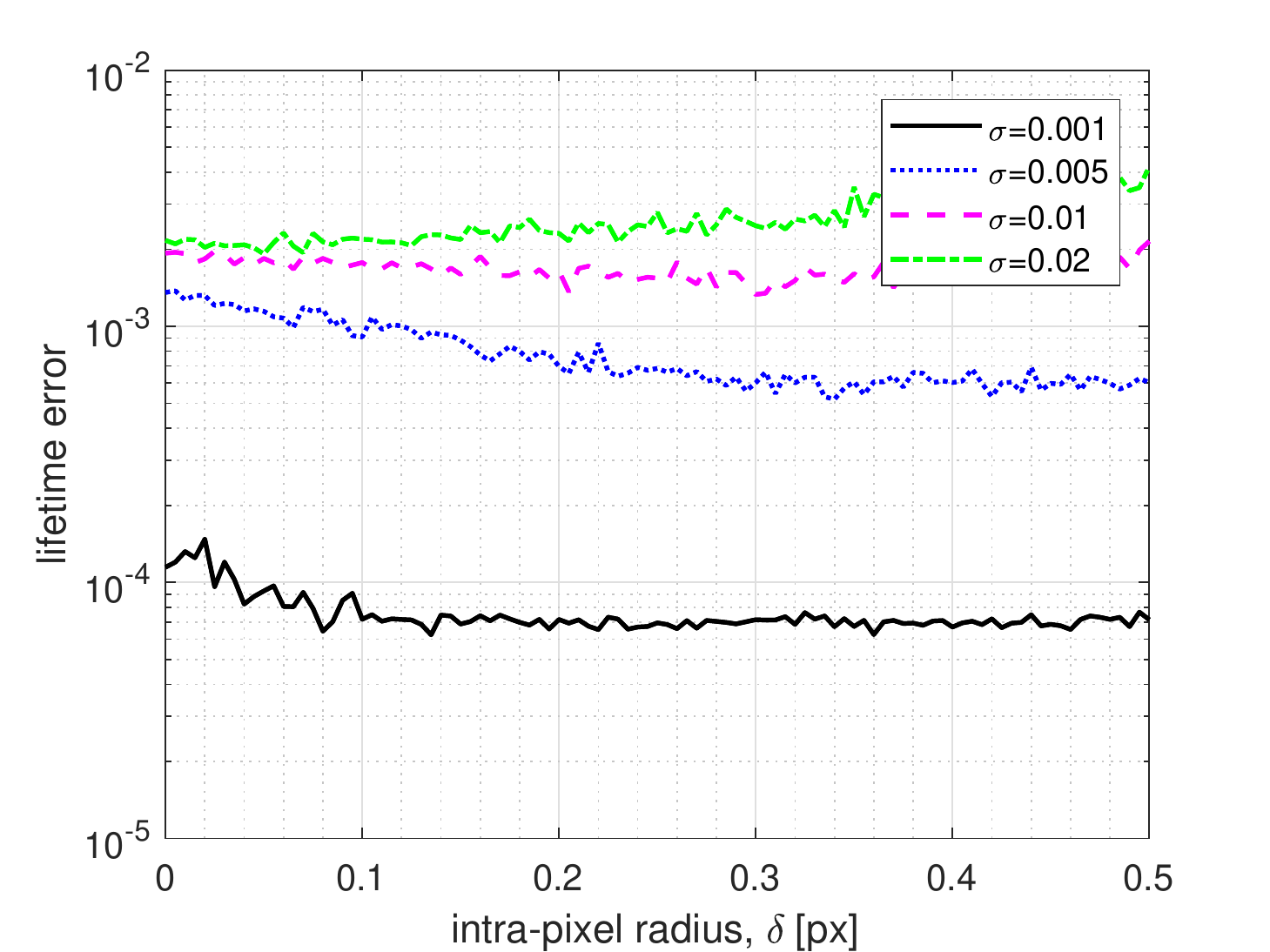} \\
    (a) w/ or w/o intra-pixel-area event & (b) various $\delta$ \\
\end{tabular}
\end{center}
    \caption{F-measurement evaluation graph with a standard deviation of data noise.}
    \label{fig:fmeasuregraph}
\end{figure}

Furthermore, we analyse the influence of the intra-pixel-area approach in terms of F-measure and lifetime error as shown in \cref{fig:fmeasuregraph}. The event data of a $5 \times 5$ window used in the analysis are manually made assuming that a single line passes, as shown in \cref{fig:floatingevent}, which is a very common condition in dataset sequence. Moreover, we add Gaussian noise in whole pixels or several randomly-chosen pixels in \cref{fig:fmeasuregraph} (a) and (b), respectively. F-measure is defined below:
\begin{equation}
    F\text{-}measure = 2\times\frac{ Recall \times Precision}{Recall + Precision},\\
\end{equation}
where recall is a fraction of determined true among a total of true, and precision is a fraction of true among a total of determined true. Also, lifetime error is the difference between the truth and the estimated values. For each evaluation, we take the average value by repeating  a total of 1,000 times.

In \cref{fig:fmeasuregraph} (a), a higher value of the intra-pixel radius, $\delta$, makes the RANSAC obtain more true inliers robustly under globally-generated noise. However, as shown in \cref{fig:fmeasuregraph} (b), it also tends to cause lifetime estimation error under scattered noise, because RANSAC may choose undesired noise data due to large intra-pixel radius. From the above analyses, $\delta$ is manually set to 0.25 for the evaluation.

\section{Experimental Evaluation}\label{sec:exp}

We qualitatively evaluate the comparison algorithms using a sequence provided by DVS ($128\times128$ pixels) in \cite{mueggler2015lifetime}. The other sequences are gray images and events captured using DVS240C ($240\times180$ pixels) for quantitative comparison of edge detection performance.

\subsection{Qualitative Evaluation}

\begin{figure}[!t]
\begin{center}
\begin{tabular}{c c}
    \includegraphics[trim={1cm 1cm 1.0cm 0cm},clip,width=0.4\textwidth]{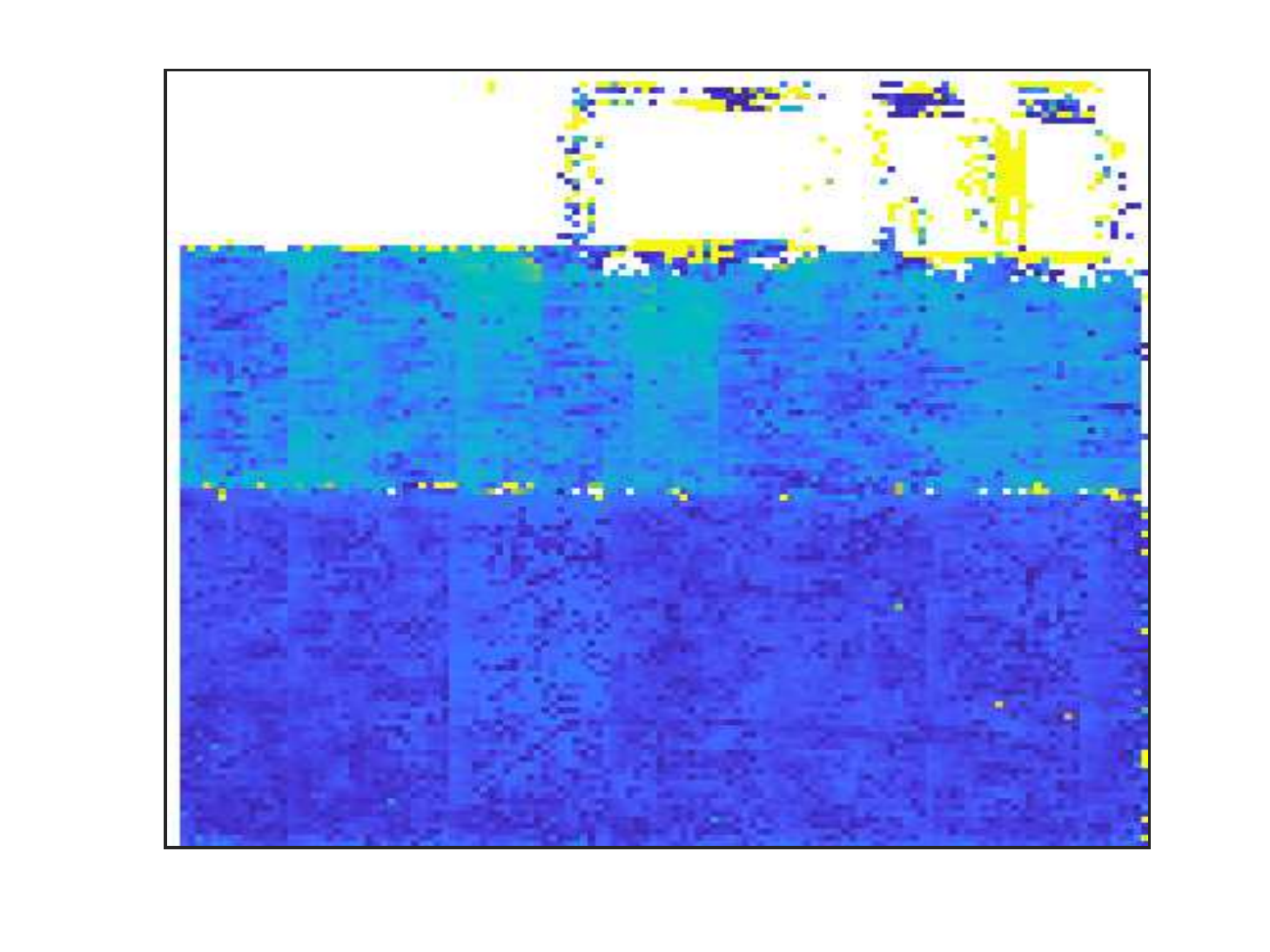} &
    \includegraphics[trim={1cm 1cm 1.0cm 0cm},clip,width=0.4\textwidth]{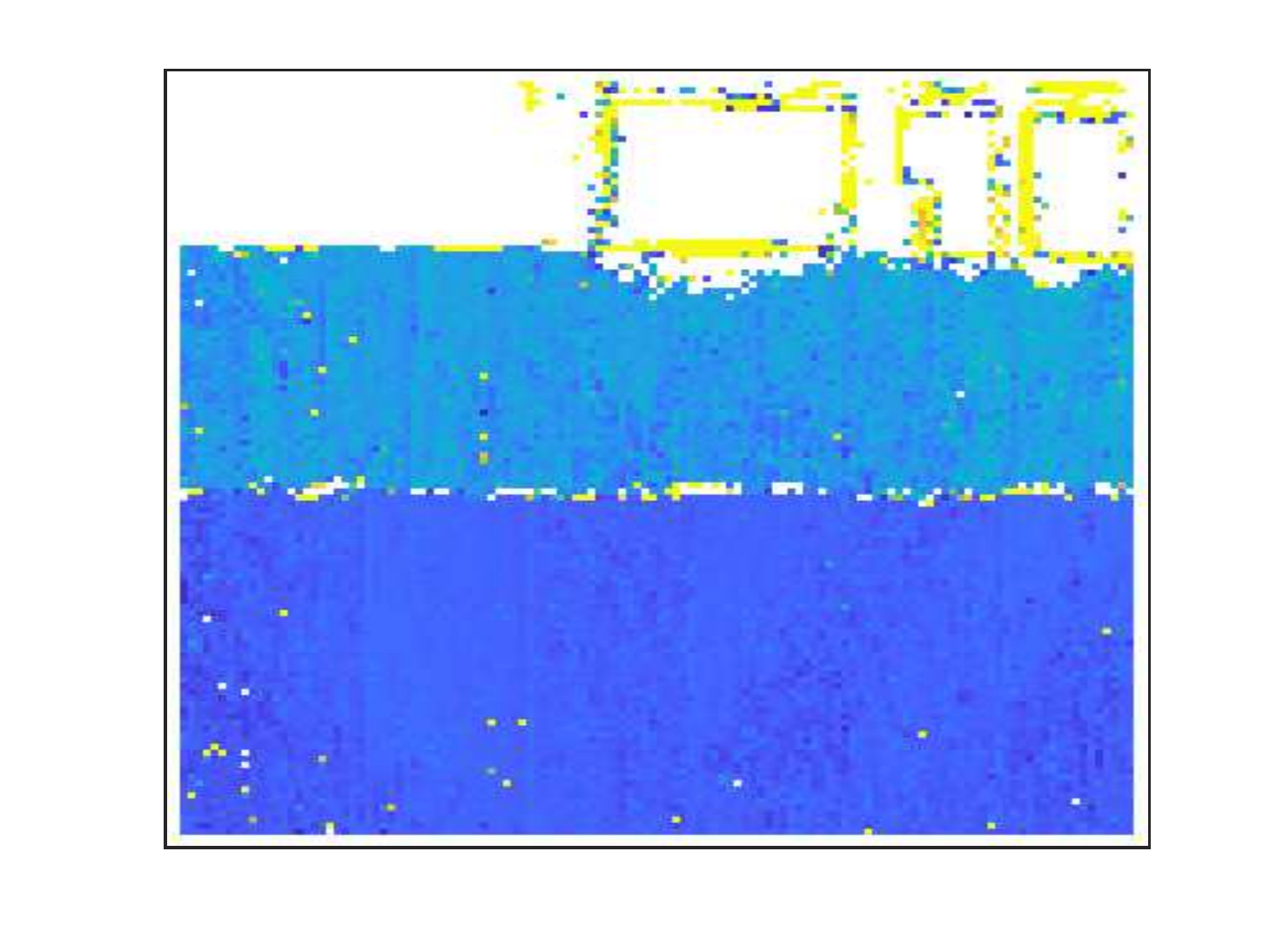} \\
    (a) E. Mueggler \etal & (b) Proposed \\
\end{tabular}
\end{center}
    \caption{The accumulation of lifetime estimates.}
    \label{fig:accumlifetime}
\end{figure}

\begin{figure}[!t]
\begin{center}
\begin{tabular}{c c}
    \includegraphics[trim={1cm 0.1cm 0.9cm 0cm},width=0.38\textwidth]{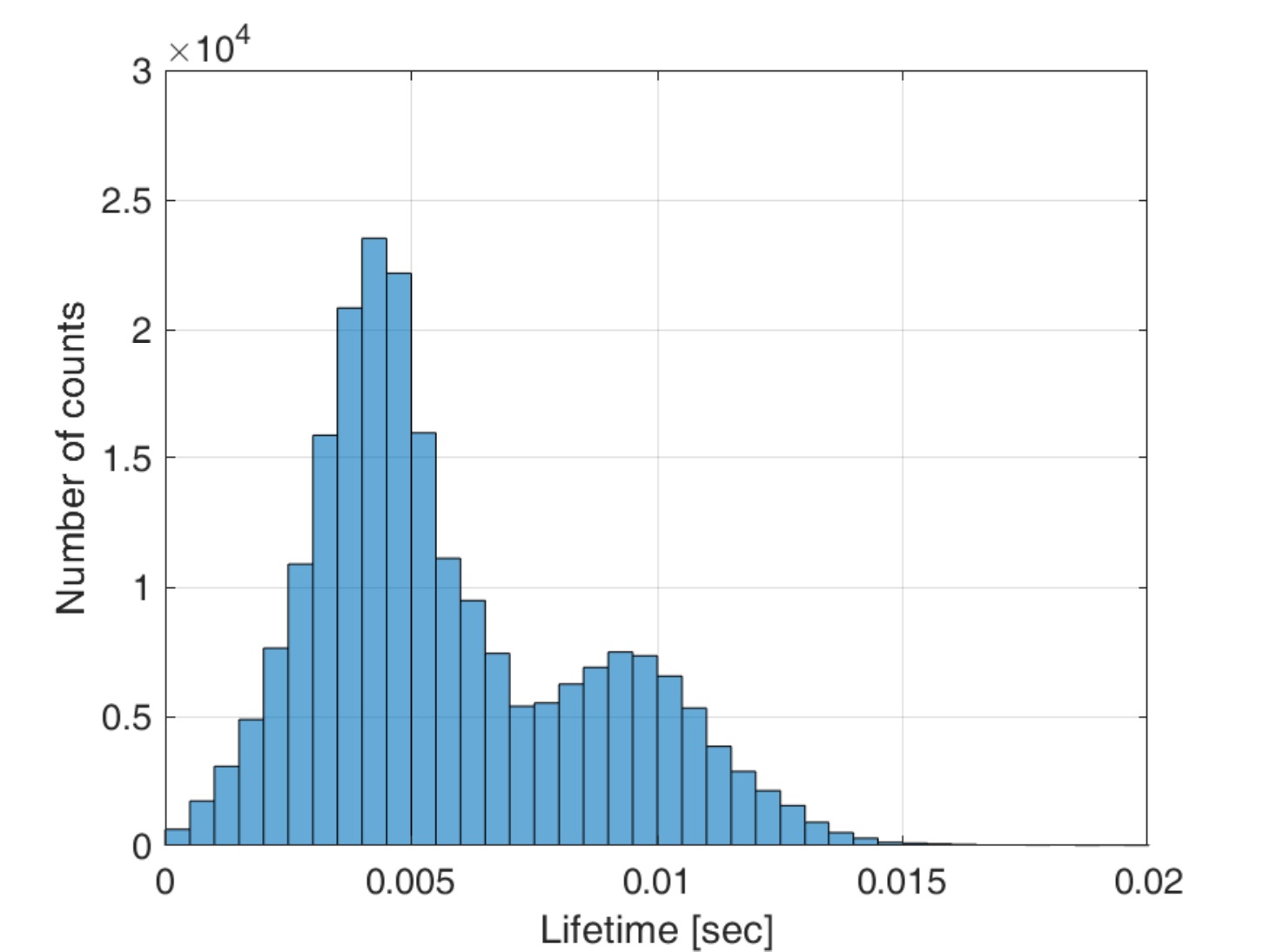} &
    \includegraphics[trim={1cm 0.1cm 0.9cm 0cm},width=0.38\textwidth]{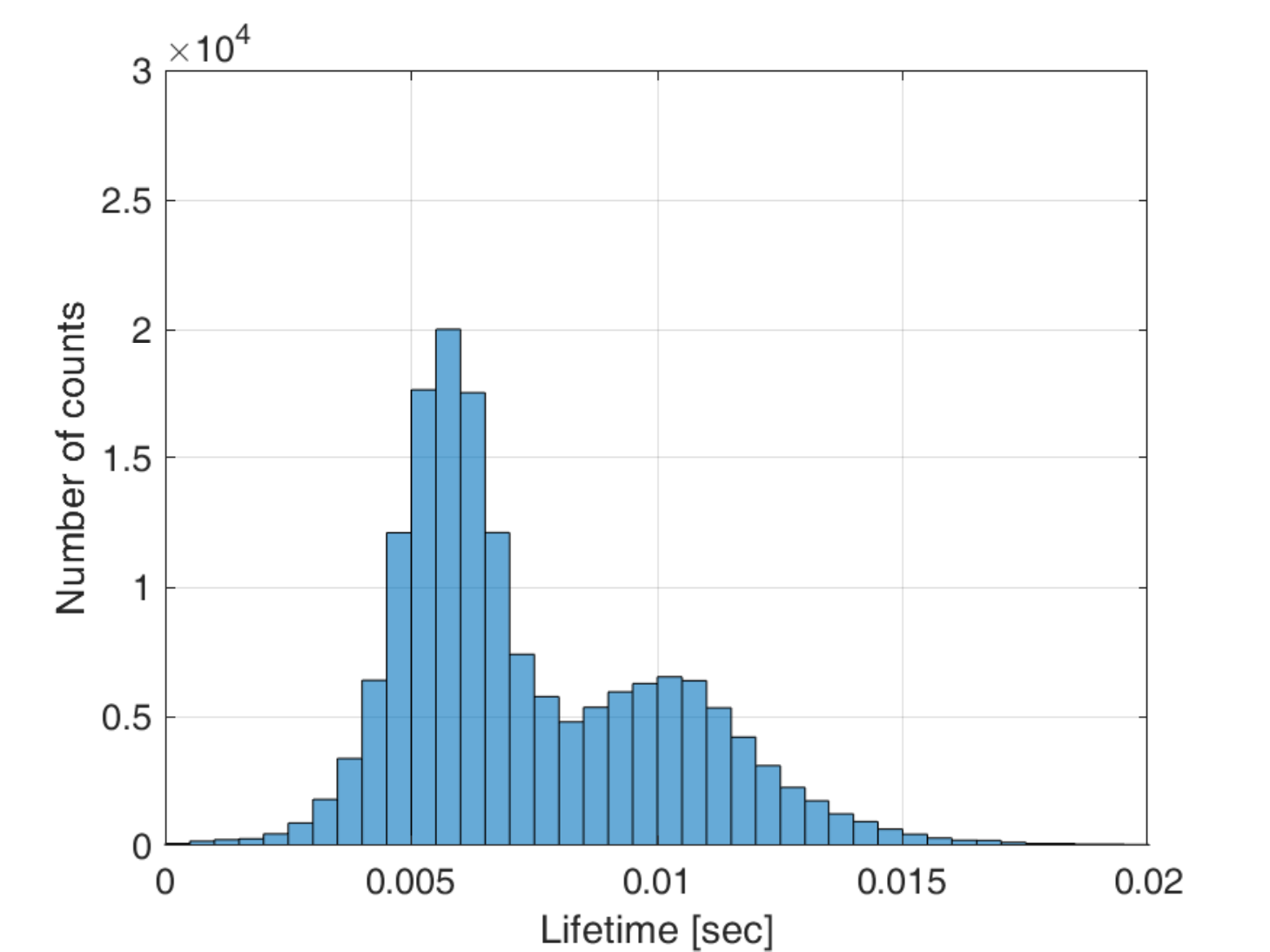} \\
    (a) E. Mueggler \etal & (b) Proposed \\
\end{tabular}
\end{center}
    \caption{The histogram of lifetime estimates.}
    \label{fig:histlifetime}
\end{figure}

For qualitative evaluation, we first show the accumulation and histogram of lifetime estimates. These analyses are provided for only a \texttt{stripe} sequence because the sequence is captured at a constant velocity making it easy to verify consistent lifetimes. Through the analysis of lifetime accumulation in \cref{fig:accumlifetime}, it is possible to see how uniform the lifetime is in areas where the stripes pass at a constant speed. Also, the analysis of lifetime histogram performed in \cite{mueggler2015lifetime} shows how precisely the lifetime is estimated through two main peaks as shown in \cref{fig:histlifetime}.

\cref{fig:accumlifetime} shows the accumulation of lifetime estimated during whole \texttt{stripes} sequence. Comparing (a) and (b), E. Mueggler \etal's algorithm shows small lifetime denoted as deep blue sometimes, while our algorithm shows uniform values across each area, in top and bottom. Further, as in the left part of the lifetime histogram in \cref{fig:histlifetime}, our result represents a sharper histogram, which means it estimates lifetime precisely.

\subsection{Quantitative Evaluation}

In this section, we evaluate our algorithm by measuring similarity to the result of the traditional edge detector from a gray image. For the similarity metric with an edge image, we use the implementation of Closest Distance Metric (CDM) in the work of \cite{bowyer1999edge,prieto2003similarity}:
\begin{equation}
    CDM_{\eta}(f,g) = 100 \hspace{.1em} \left(1- \frac{\mathcal{C}(\mathcal{M}_{cd}(f,g))}{|f \cup g|}\right),
\end{equation}
where $\eta$ is the neighbourhood radius to find matching edge pixels between two images, $f$ and $g$, $\mathcal{C}(\mathcal{M}_{cd}(f,g))$ is the cost of a pair matched by closest-distance criteria, and $|f \cup g|$ is the number of edge pixels belonging to $f$ or $g$. In evaluation, $\eta$ is set to 3, and we use 8-connected chessboard distance metric as mentioned in \cite{prieto2003similarity}. In addition, for comparison, the results of the Canny Edge Detector\cite{canny1986computational} are considered as groundtruth and the performance of algorithms is confirmed by various Canny Edge Detector parameters.

\begin{figure}[!t]
\begin{center}
\begin{tabular}{c c c c}
    \includegraphics[width=0.22\textwidth]{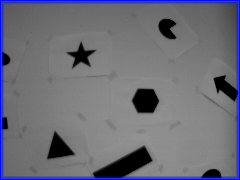} & 
    \includegraphics[width=0.22\textwidth]{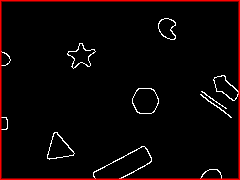} & 
    \includegraphics[width=0.22\textwidth]{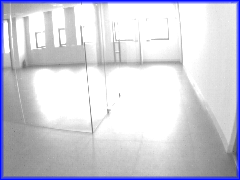} & 
    \includegraphics[width=0.22\textwidth]{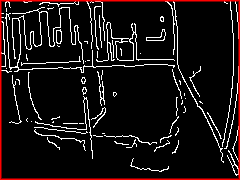} \\
    \includegraphics[width=0.22\textwidth]{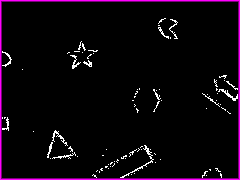} &
    \includegraphics[width=0.22\textwidth]{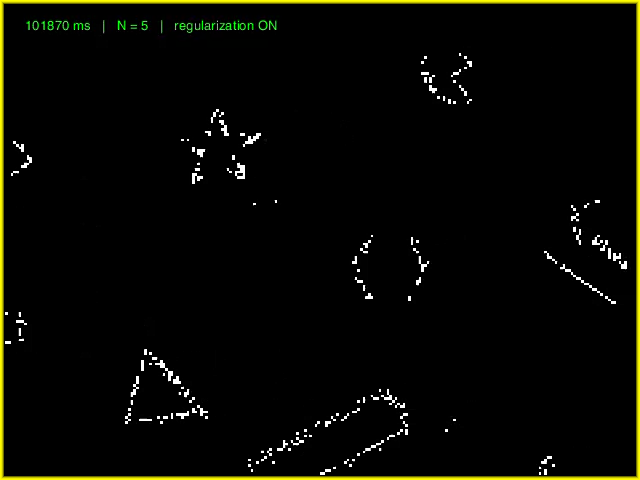} &
    \includegraphics[width=0.22\textwidth]{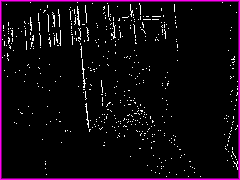} &
    \includegraphics[width=0.22\textwidth]{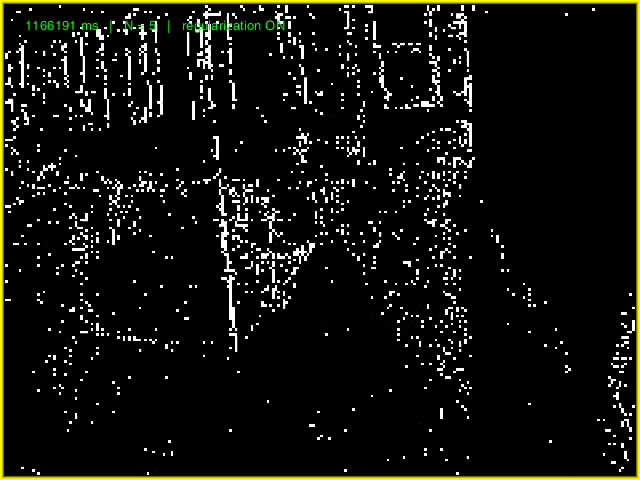} \\
    \includegraphics[width=0.22\textwidth]{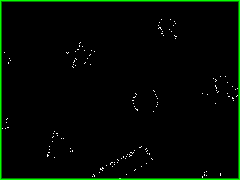} &
    \includegraphics[width=0.22\textwidth]{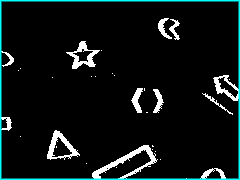} &
    \includegraphics[width=0.22\textwidth]{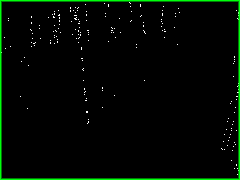} &
    \includegraphics[width=0.22\textwidth]{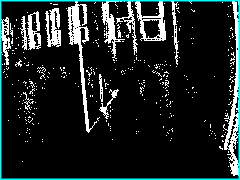} \\
    \multicolumn{2}{c}{(a) \texttt{shape\_translation} sequence} & \multicolumn{2}{c}{(b) \texttt{corridor} sequence (self-collected)} \\
\end{tabular}
\end{center}
    \caption{The result of : gray image (blue), groundtruth computed by Canny Edge Detector (red), E. Mueggler \etal 's algorithm (yellow), 30ms (cyan), 1ms (green) accumulation, and proposed algorithm (magenta) for each sequence in clockwise from top-left.}
    \label{fig:result}
    \vspace{-2mm}
\end{figure}

\begin{figure}[!t]
\begin{center}
\begin{tabular}{c c}
    \multicolumn{2}{c}{\includegraphics[trim={1.2cm 0cm 0cm 0cm},clip,width=\textwidth]{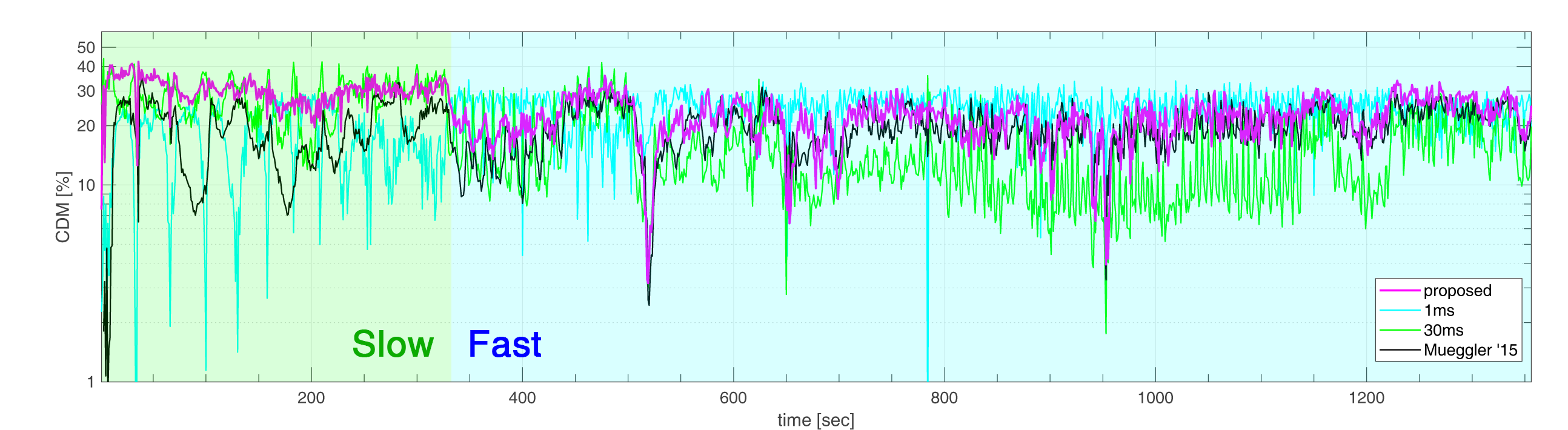}} \\
    \multicolumn{2}{c}{(a) Similarity over time}\\
    \includegraphics[width=0.48\textwidth]{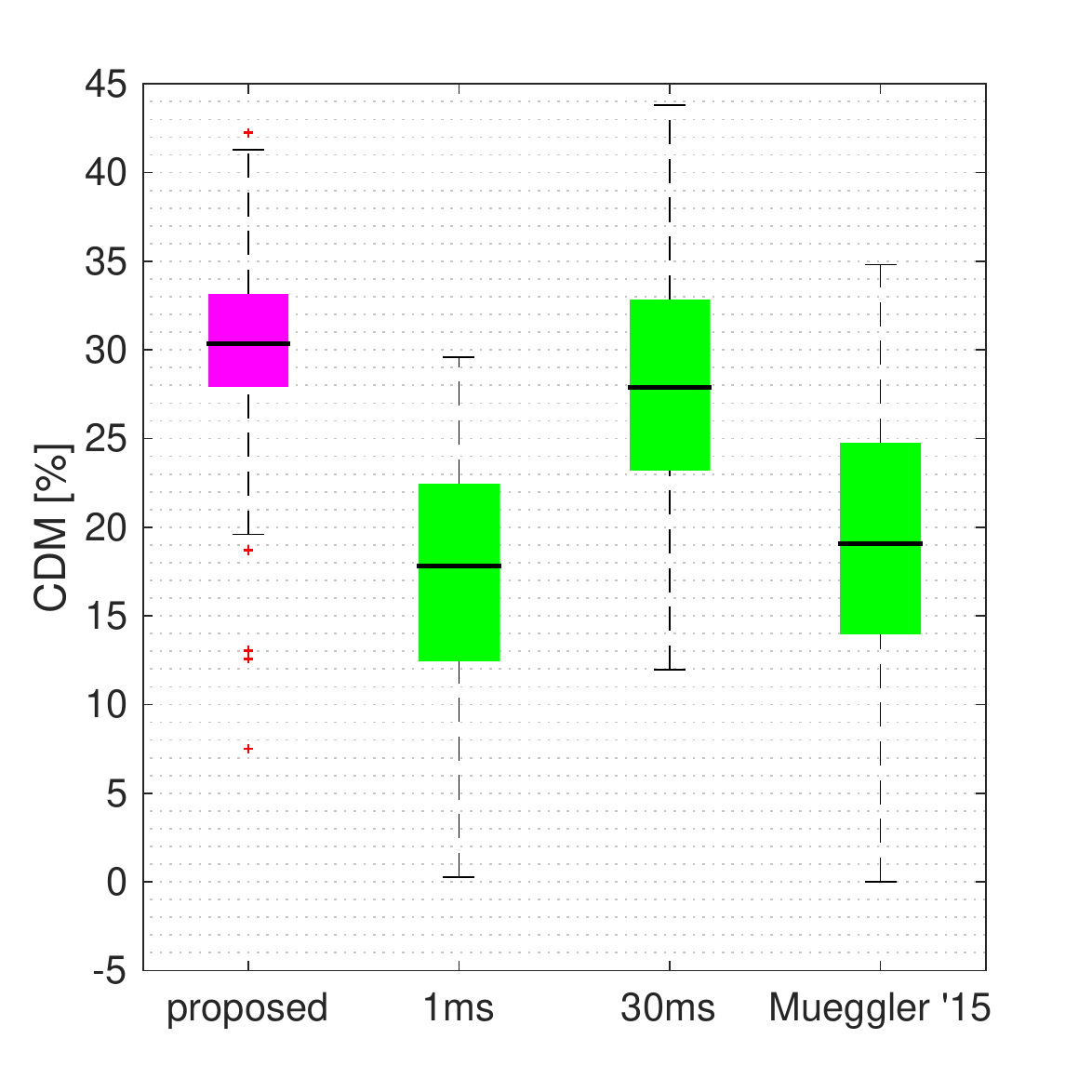} &
    \includegraphics[width=0.48\textwidth]{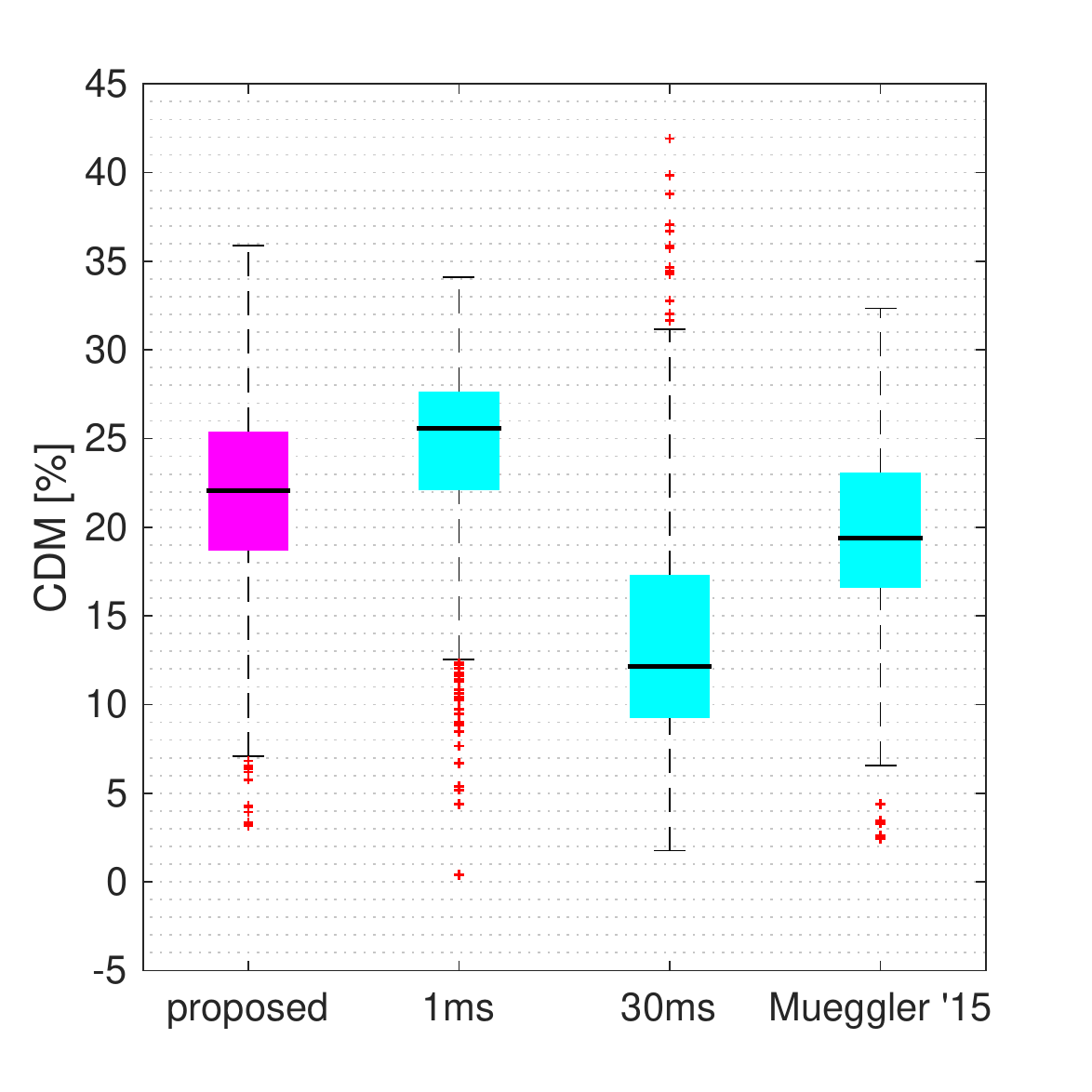} \\
    (b) Boxplot on the slow section & (c) Boxplot on the fast section\\
\end{tabular}
\end{center}
    \caption{Performance analysis of the whole sequence on the \texttt{shapes\_translation} sequence. The speed of camera is slow in the front and fast in the back. Thus, 1ms and 30ms accumulation of event shows different performance depending on the camera's speed.}
    \label{fig:boxplot}
\end{figure}

\cref{fig:result} shows the result of: gray image, Canny edge, proposed algorithm, 1ms, 30ms accumulation, and E. Mueggler\cite{mueggler2015lifetime}. Although the lifetime estimation including ours and E. Mueggler's algorithm tend to leave tracks on the edge image, they show a thinner edge than the result of accumulation. In particular, the performance of edge detection over time for a whole sequence is shown in \cref{fig:boxplot}. Our algorithm (magenta) shows relatively consistent performance in terms of the CDM rather the other methods regardless of the camera motion as shown in \cref{fig:boxplot} (b) and (c). Sometimes, our algorithm shows bad performance denoted as red-cross outliers when the camera motion is almost stationary. Note that the performances of the accumulation methods depend on the speed of the camera in the bottom figure.

\section{Conclusion}\label{sec:con}

In the work, we proposed an algorithm to detect an edge from events of a dynamic vision sensor by virtue of the lifetime estimation with an event buffer and intra-pixel-area approach. The designed event buffer regards isolated events as noise and reduces them effectively. Also, the proposed intra-pixel-area approach makes the algorithm find local plane fitted on SAE robustly so that the lifetime is estimated precisely. Moreover, we analysed the effectiveness of the intra-pixel-area approach by the F-measure versus the standard deviation of timestamp noise and the estimation error versus the intra-pixel radius. For evaluation, we  qualitatively compare our algorithm with the existing lifetime estimation or na\"ive event accumulation approach. In addition, with the well-known edge similarity metric, we measure the performance of the algorithm quantitatively. Then, we confirm that our algorithm performs better in terms of sharpness and similarity to the Canny edge than the accumulation of events over fixed counts or time intervals, and the existing lifetime estimation algorithm. Further, by utilising the detected edge, the proposed algorithm can be employed as a pre-processing part of edge-based visual odometry or SLAM for autonomous robots.

\bibliography{egbib}

\begin{thebibliography}{21}
\providecommand{\natexlab}[1]{#1}
\providecommand{\url}[1]{\texttt{#1}}
\expandafter\ifx\csname urlstyle\endcsname\relax
  \providecommand{\doi}[1]{doi: #1}\else
  \providecommand{\doi}{doi: \begingroup \urlstyle{rm}\Url}\fi

\bibitem[Barranco et~al.(2015)Barranco, Teo, Fermuller, and
  Aloimonos]{barranco2015contour}
Francisco Barranco, Ching~L Teo, Cornelia Fermuller, and Yiannis Aloimonos.
\newblock Contour detection and characterization for asynchronous event
  sensors.
\newblock In \emph{Proceedings of the IEEE International Conference on Computer
  Vision}, pages 486--494, 2015.

\bibitem[Benosman et~al.(2014)Benosman, Clercq, Lagorce, Ieng, and
  Bartolozzi]{benosman2014event}
Ryad Benosman, Charles Clercq, Xavier Lagorce, Sio-Hoi Ieng, and Chiara
  Bartolozzi.
\newblock Event-based visual flow.
\newblock \emph{IEEE Trans. Neural Netw. Learning Syst.}, 25\penalty0
  (2):\penalty0 407--417, 2014.

\bibitem[Bowyer et~al.(1999)Bowyer, Kranenburg, and Dougherty]{bowyer1999edge}
K~Bowyer, C~Kranenburg, and S~Dougherty.
\newblock Edge detector evaluation using empirical roc curves.
\newblock In \emph{Proceedings. 1999 IEEE Computer Society Conference on
  Computer Vision and Pattern Recognition (Cat. No PR00149)}, volume~1, pages
  354--359. IEEE, 1999.

\bibitem[Brandli et~al.(2014)Brandli, Berner, Yang, Liu, and
  Delbruck]{brandli2014240}
Christian Brandli, Raphael Berner, Minhao Yang, Shih-Chii Liu, and Tobi
  Delbruck.
\newblock A 240$\times$ 180 130 db 3 $\mu$s latency global shutter
  spatiotemporal vision sensor.
\newblock \emph{IEEE Journal of Solid-State Circuits}, 49\penalty0
  (10):\penalty0 2333--2341, 2014.

\bibitem[Br{\"a}ndli et~al.(2016)Br{\"a}ndli, Strubel, Keller, Scaramuzza, and
  Delbruck]{brandli2016elised}
Christian Br{\"a}ndli, Jonas Strubel, Susanne Keller, Davide Scaramuzza, and
  Tobi Delbruck.
\newblock Elised—an event-based line segment detector.
\newblock In \emph{2016 Second International Conference on Event-based Control,
  Communication, and Signal Processing (EBCCSP)}, pages 1--7. IEEE, 2016.

\bibitem[Canny(1986)]{canny1986computational}
John Canny.
\newblock A computational approach to edge detection.
\newblock \emph{IEEE Transactions on Pattern Analysis and Machine
  Intelligence}, 8\penalty0 (6):\penalty0 679--698, 1986.

\bibitem[Censi and Scaramuzza(2014)]{censi2014low}
Andrea Censi and Davide Scaramuzza.
\newblock Low-latency event-based visual odometry.
\newblock In \emph{Robotics and Automation (ICRA), 2014 IEEE International
  Conference on}, pages 703--710. IEEE, 2014.

\bibitem[Gallego et~al.(2018)Gallego, Rebecq, and
  Scaramuzza]{gallego2018unifying}
Guillermo Gallego, Henri Rebecq, and Davide Scaramuzza.
\newblock A unifying contrast maximization framework for event cameras, with
  applications to motion, depth, and optical flow estimation.
\newblock In \emph{IEEE Int. Conf. Comput. Vis. Pattern Recog.(CVPR)},
  volume~1, 2018.

\bibitem[Jose~Tarrio and Pedre(2015)]{jose2015realtime}
Juan Jose~Tarrio and Sol Pedre.
\newblock Realtime edge-based visual odometry for a monocular camera.
\newblock In \emph{Proceedings of the IEEE International Conference on Computer
  Vision}, pages 702--710, 2015.

\bibitem[Kim et~al.(2018)Kim, Kim, Lee, and Kim]{kim2018edge}
Changhyeon Kim, Pyojin Kim, Sangil Lee, and H~Jin Kim.
\newblock Edge-based robust rgb-d visual odometry using 2-d edge divergence
  minimization.
\newblock In \emph{2018 IEEE/RSJ International Conference on Intelligent Robots
  and Systems (IROS)}, pages 1--9. IEEE, 2018.

\bibitem[Kim et~al.(2008)Kim, Handa, Benosman, Ieng, and
  Davison]{kim2008simultaneous}
Hanme Kim, Ankur Handa, Ryad Benosman, Sio-Hoi Ieng, and Andrew~J Davison.
\newblock Simultaneous mosaicing and tracking with an event camera.
\newblock \emph{J. Solid State Circ}, 43:\penalty0 566--576, 2008.

\bibitem[Kim et~al.(2016)Kim, Leutenegger, and Davison]{kim2016real}
Hanme Kim, Stefan Leutenegger, and Andrew~J Davison.
\newblock Real-time 3d reconstruction and 6-dof tracking with an event camera.
\newblock In \emph{European Conference on Computer Vision}, pages 349--364.
  Springer, 2016.

\bibitem[Lichtsteiner et~al.(2008)Lichtsteiner, Posch, and
  Delbruck]{lichtsteiner2008128}
Patrick Lichtsteiner, Christoph Posch, and Tobi Delbruck.
\newblock A 128$\times $128 120 db 15$\mu $ s latency asynchronous temporal
  contrast vision sensor.
\newblock \emph{IEEE journal of solid-state circuits}, 43\penalty0
  (2):\penalty0 566--576, 2008.

\bibitem[Maity et~al.(2017)Maity, Saha, and Bhowmick]{maity2017edge}
Soumyadip Maity, Arindam Saha, and Brojeshwar Bhowmick.
\newblock Edge slam: Edge points based monocular visual slam.
\newblock In \emph{Proceedings of the IEEE International Conference on Computer
  Vision}, pages 2408--2417, 2017.

\bibitem[Mueggler et~al.(2014)Mueggler, Huber, and
  Scaramuzza]{mueggler2014event}
Elias Mueggler, Basil Huber, and Davide Scaramuzza.
\newblock Event-based, 6-dof pose tracking for high-speed maneuvers.
\newblock In \emph{Intelligent Robots and Systems (IROS 2014), 2014 IEEE/RSJ
  International Conference on}, pages 2761--2768. IEEE, 2014.

\bibitem[Mueggler et~al.(2015)Mueggler, Forster, Baumli, Gallego, and
  Scaramuzza]{mueggler2015lifetime}
Elias Mueggler, Christian Forster, Nathan Baumli, Guillermo Gallego, and Davide
  Scaramuzza.
\newblock Lifetime estimation of events from dynamic vision sensors.
\newblock In \emph{2015 IEEE international conference on Robotics and
  Automation (ICRA)}, pages 4874--4881. IEEE, 2015.

\bibitem[Mueggler et~al.(2017)Mueggler, Bartolozzi, and
  Scaramuzza]{mueggler2017fast}
Elias Mueggler, Chiara Bartolozzi, and Davide Scaramuzza.
\newblock Fast event-based corner detection.
\newblock In \emph{28th British Machine Vision Conference (BMVC)}, 2017.

\bibitem[Prieto and Allen(2003)]{prieto2003similarity}
Miguel~Segui Prieto and Alastair~R Allen.
\newblock A similarity metric for edge images.
\newblock \emph{IEEE Transactions on Pattern Analysis and Machine
  Intelligence}, 25\penalty0 (10):\penalty0 1265--1273, 2003.

\bibitem[Seifozzakerini et~al.(2016)Seifozzakerini, Yau, Zhao, and
  Mao]{seifozzakerini2016event}
Sajjad Seifozzakerini, Wei-Yun Yau, Bo~Zhao, and Kezhi Mao.
\newblock Event-based hough transform in a spiking neural network for multiple
  line detection and tracking using a dynamic vision sensor.
\newblock In \emph{BMVC}, 2016.

\bibitem[Zhou et~al.(2019)Zhou, Li, and Kneip]{zhou2019canny}
Yi~Zhou, Hongdong Li, and Laurent Kneip.
\newblock Canny-vo: Visual odometry with rgb-d cameras based on geometric
  3-d--2-d edge alignment.
\newblock \emph{IEEE Transactions on Robotics}, 35\penalty0 (1):\penalty0
  184--199, 2019.

\bibitem[Zhu et~al.(2017)Zhu, Atanasov, and Daniilidis]{zhu2017event}
Alex~Zihao Zhu, Nikolay Atanasov, and Kostas Daniilidis.
\newblock Event-based visual inertial odometry.
\newblock In \emph{CVPR}, pages 5816--5824, 2017.

\end{thebibliography}
\end{document}